\documentclass{article}
\usepackage{ijcai25}

\usepackage{times}
\usepackage{soul}
\usepackage{url}
\usepackage[hidelinks]{hyperref}
\usepackage[utf8]{inputenc}
\usepackage[small]{caption}
\usepackage{graphicx}
\usepackage{amsmath}
\usepackage{amsthm}
\usepackage{booktabs}
\usepackage{algorithm}
\usepackage{algorithmic}
\usepackage[switch]{lineno}
\usepackage{bbding}
\usepackage{enumitem}
\usepackage{multirow}
\usepackage{bbding}
\usepackage{CJK}
\setenumerate[1]{itemsep=0pt,partopsep=0pt,parsep=\parskip,topsep=0pt}
\setitemize[1]{itemsep=0pt,partopsep=0pt,parsep=\parskip,topsep=0pt}

\title{Multimodal Fake News Detection: MFND Dataset and Shallow-Deep Multitask Learning}

\author{
Ye Zhu$^1$
\and
Yunan Wang$^1$\And
Zitong Yu$^{2,3,4}$\thanks{Corresponding Author}\\
\affiliations
$^1$School of Artificial Intelligence, Hebei University of Technology\\
$^2$School of Computing and Information Technology, Great Bay University\\
$^3$Guangdong Provincial Key Laboratory of Intelligent Information Processing \& Shenzhen Key Laboratory of Media Security, Shenzhen University\\
$^4$Dongguan Key Laboratory for Intelligence and Information Technology\\
}

\begin{document}
\maketitle

\begin{abstract}
    Multimodal news contains a wealth of information and is easily affected by deepfake modeling attacks. To combat the latest image and text generation methods, we present a new Multimodal Fake News Detection dataset (MFND) containing 11 manipulated types, designed to detect and localize highly authentic fake news. Furthermore, we propose a Shallow-Deep Multitask Learning (SDML) model for fake news, which fully uses unimodal and mutual modal features to mine the intrinsic semantics of news. Under shallow inference, we propose the momentum distillation-based light punishment contrastive learning for fine-grained uniform spatial image and text semantic alignment, and an adaptive cross-modal fusion module to enhance mutual modal features. Under deep inference, we design a two-branch framework to augment the image and text unimodal features, respectively merging with mutual modalities features, for four predictions via dedicated detection and localization projections. Experiments on both mainstream and our proposed datasets demonstrate the superiority of the model. Codes and dataset are released at https://github.com/yunan-wang33/sdml.
\end{abstract}

\section{Introduction}
Multimodal news composed of visual and textual components has become the mainstream information dissemination method. However, the malicious abuse of Large Language Models (LLMs) ~\cite{kenton:bert,radford:language} and Deep Generative Models ~\cite{dolhansky:deepfake,zhang:genface} are challenging media credibility, brought lots of negative impacts to society. Computer Vision (CV) and Natural Language Processing (NLP) fields have respectively proposed many approaches ~\cite{xie:fusionmamba,gao:mini}, but most of these do not consider news as a whole, or provide inaccurate analysis of semantics.
\nocite{kenton:bert}
\nocite{radford:language}
\nocite{dolhansky:deepfake}
\nocite{zhang:genface}
\nocite{xie:fusionmamba}
\nocite{gao:mini}

\begin{figure}[!t]
    \centering
    \setlength{\abovecaptionskip}{0.2cm}
    \setlength{\belowcaptionskip}{-0.5cm}
    {\includegraphics{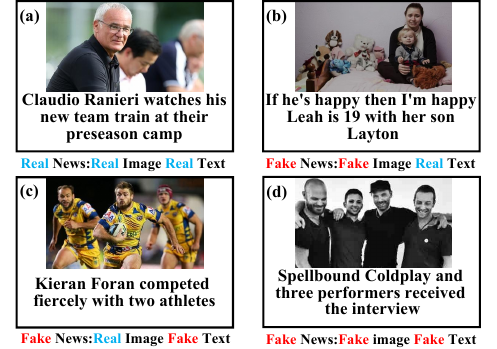}}%
    \caption{Illustrates of the news from the MFND dataset. (a) Real News of Real Image Real Text, (b) Fake News of Fake Image Real Text, (c) Fake News of Real Image Fake Text, (d) Fake News of Fake Image Fake Text.}
    \label{fig1}
\end{figure}

Multimodal fake news detection research can be outlined into two categories: traditional learning methods and deep learning methods. Traditional methods ~\cite{Liu:Rumor,Ma:websites} require handcrafted features with expert knowledge, and are inefficient in growing data-driven fashion. Recent studies use deep learning methods ~\cite{wang:style,luvembe:caf} which are based on neural networks to capture features, but most of them are limited to binary detection. From a multimodal perspective, fake news requires multitasking for detection and localization to mine deep inference information, Fig.~\ref{fig1} illustrates four types of news from the MFND dataset, where the semantic similarity values of true and fake news are alike, and it is difficult to reason effectively only through the shallow binary categorization task with mutual features.
\nocite{Liu:Rumor}
\nocite{Ma:websites}
\nocite{wang:style}
\nocite{luvembe:caf}

\begin{figure*}[!t]
    \centering
    \setlength{\abovecaptionskip}{0.2cm}
    \setlength{\belowcaptionskip}{-0.5cm}
        \begin{minipage}[b]{0.25\linewidth}
            \centering
            {\includegraphics{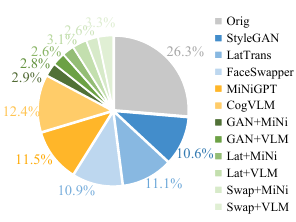}}
            \caption*{(a)}
        \end{minipage}
        \begin{minipage}[b]{0.7\linewidth}
            \setlength{\tabcolsep}{2pt}
            \setlength{\abovecaptionskip}{0.2cm}
            \centering
            \begin{tabular}{cccccccccc}
                \toprule
                Datasets & News & Image & Text & Deepfake & Forgery & Real & Fake & Year \\
                \midrule
                Twitter & \Checkmark & \XSolidBrush & \XSolidBrush & \XSolidBrush & \XSolidBrush  & 6026 & 7898 & 2017\\
                Pheme & \Checkmark & \XSolidBrush & \XSolidBrush & \XSolidBrush & \XSolidBrush & 3820 & 1972 & 2018\\
                Weibo & \Checkmark & \XSolidBrush & \XSolidBrush & \XSolidBrush & \XSolidBrush & 4640 & 4488 & 2020\\
                NewsCLIPpings & \Checkmark & \XSolidBrush & \XSolidBrush & \XSolidBrush & \XSolidBrush  & 85360 & 85360 & 2021\\
                DGM$^4$ & \Checkmark & \Checkmark & \Checkmark & GAN & 8 & 77426 & 152574 & 2023\\
                MFND & \Checkmark & \Checkmark & \Checkmark & GAN+LLM & 11 & 32869 & 92131 & 2025\\
                \bottomrule
            \end{tabular}
            \caption*{(b)}
        \end{minipage}
    \caption{Statistics of (a) our proposed MFND dataset, and (b) comparison with 5 other multimodal fake news datasets.}
    \label{fig2}
\end{figure*}

To solve the above problems, we propose a Shallow-Deep Multitask Learning (SDML) model for fake news detection and localization, which exports binary prediction for media news, and detects manipulated images and text meanwhile forecasting image bounding boxes. To solve the problem of over-penalization of hard negative image-text pair in feature alignment, we propose a light punishment contrastive learning based on momentum distillation. Moreover, we propose adaptive cross-modal fusion to blend the aligned image and text unimodal features and spontaneously adjust the weight between the two modalities. Finally, we leverage the enhanced unimodal and multimodal features to predict images and text detection and localization results in different branches. To counter the more truthful AI fake news, we establish the MFND dataset which is about media news with humans as the principal part. The proposed MFND dataset contains four multi-modality types and employs post-processing to simulate the real scene, providing media news true-false binary labels, manipulated image labels, manipulated text labels, and manipulated image localization labels.

Overall, the main contributions of this paper are as:
\begin{itemize}
    \setlength{\itemsep}{0.5pt}
    \setlength{\parsep}{0.5pt}
    \setlength{\parskip}{0pt}
    \item We contribute the MFND dataset which uses 11 state-of-the-art image and text manipulation methods and provides rich detection and localization labels that fit a wide range of realistic scenarios.
    \item We propose a Shallow-Deep Multitask Learning (SDML) model for fake news detection and localization, which fuses image and text features after alignment, combining mutual modality with augmented unimodality for fine-grained semantic inference.
    \item The proposed SDML model achieves state-of-the-art detection and localization performance on four benchmark datasets under both multi-modal multi-task and multi-modal single-task settings.
\end{itemize}

\section{Related Work}
\subsection{Deepfake Detection}
Current image deepfake researches are mostly based on spatial and frequency domains, such as texture features, noise distribution, blending artifacts, and so on. Zhao ~\cite{zhao:multi} applied multi-attention to obtain image local details and aggregate low-level and high-level texture features. Shi ~\cite{shi:transformer} proposed stacked multi-scale transformers to mine image structural anomalies on blocks of different sizes. Some recent methods incorporate the frequency domain information as data enhancement, Gao ~\cite{Gao:TBNet} using the frequency and spatial domains as dual streams to locate results in a hierarchical fusion manner. Li ~\cite{li:edge} fused graph convolutional processing features to predict image forgery binary masks guided by dual attention. However, none of them combine with textual modalities whose form have more widespread and harmful.
\nocite{zhao:multi}
\nocite{shi:transformer}
\nocite{Gao:TBNet}
\nocite{li:edge}

\subsection{Multimodal Fake News Detection}
Several works have investigated multimodal fake news detection. Qian ~\cite{qian:hierarchical} encodes features then hierarchically fuse and sends them to decode getting news results. Zhang ~\cite{zhang:cross} introduces cross-contrastive learning and an attention mechanism guides reasoning. Ma ~\cite{ma:event} integrates inconsistencies at the event level and utilizes graph capture plausibility for robust predictions. These works are limited to binary detection, and datasets are miniature while fake news structures are mostly image-text exchanged without deepfake technology. Shao ~\cite{shao:detecting} defined detection and localization tasks for the first time and built a dataset containing deepfake forgery. However, the types of manipulation are not rich enough. In addition, due to the complexity of the similarity relationship between different texts, the strict text localization labels hinder capturing the accurate image-text depth semantics, and localization annotation also requires extra computational cost. In this paper, we redefine the task of multimodal fake news detection and localization and build a dataset containing more deepfake news with a small labeling cost.
\nocite{qian:hierarchical}
\nocite{zhang:cross}
\nocite{ma:event}
\nocite{shao:detecting}

\section{MFND Dataset}
Most established fake multimodal datasets usually adopt coarse-grained annotations for binary detection and collect hand-generated or contextual semantic outer pairs. To facilitate the study of multimodal fake news detection, the large-scale and diverse Multimodal Fake News Detection (MFND) dataset is introduced, which covers keyword and sentiment reversal, summary induction, and keyword substitution manipulation techniques. Specifically, MFND is built on the original VisualNews dataset ~\cite{liu:visual}, which contains numerous real social news. We select image-text pairs centered on humans as the source pool $O=\left\{p_{o} \mid p_{o}=\left(I_{o}, T_{0}\right)\right\}$ through data filtering.
\nocite{liu:visual}

\begin{figure*}[!t]
    \centering
    \setlength{\abovecaptionskip}{0.2cm}
    \setlength{\belowcaptionskip}{-0.5cm}
    \includegraphics{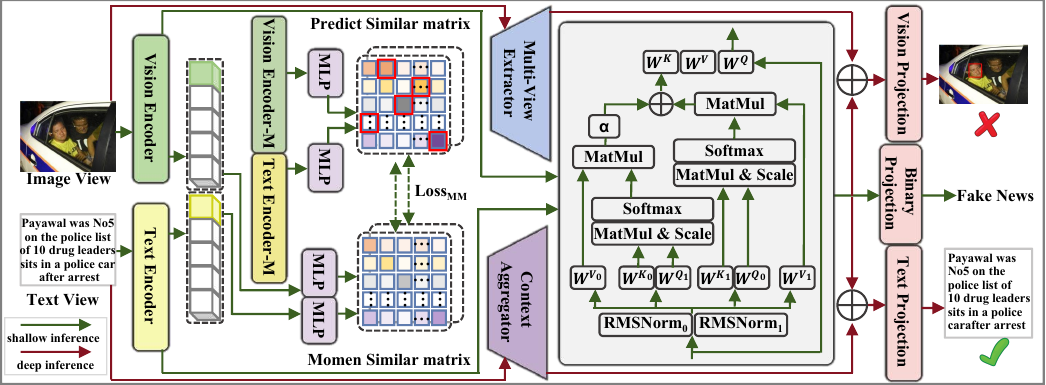}
    \caption{Illustration of the proposed Shallow-Deep Multitask Learning (SDML) method. As for the shallow inference with green lines, we encode two single modalities using different pre-trained Encoders, align the embeddings by contrastive learning, and obtain mutual modality after adaptive fusion for media news binary classification. As for the deep inference with red lines, we enhance features from image and text modalities in a two-branch framework, combined with the fusion feature for detection and localization.} 
    \label{fig3}
\end{figure*}

\subsection{Data Collection}
Deepfake generates fake faces based on deep learning methods mainly divided into three categories, i.e., Entire Face Synthesis (EFS) ~\cite{karras:alias,Ho:denos}, Attribute Manipulation (AM) ~\cite{Yao:latent,pernuvs:maskfacegan}, and Face Swap (FS) ~\cite{ma:learning,xu:high}. EFS uses generative techniques to obtain non-existent fake faces from random noise, AM edits the facial attributes of the original image to produce new forged faces through deep learning models, and FS utilizes neural networks to replace the face of the source image with the face of the target image. We create the MFND dataset through the above forgery techniques.
\nocite{karras:alias}
\nocite{Ho:denos}
\nocite{Yao:latent}
\nocite{pernuvs:maskfacegan}
\nocite{ma:learning}
\nocite{xu:high}

\subsubsection{Real Image Real Text}
The source pool $O$ after filtering the VisualNews dataset contains 200k valid image-text pairs, we randomly draw some sample pairs from it as the initial part of the MFND dataset, given an example as Fig.~\ref{fig1}(a). We record binary labels of the media news, images, and texts respectively, constantly updating the source pool to remove the drawn sample pairs.

\subsubsection{Fake Image Real text}
We subject the sampled pairs with an equal probability to EFS, AM, and FS manipulation. Specifically utilizing CelebAHQ ~\cite{karras:progressive} pre-trained StyleGAN3 changes the main faces in the image to noise-generated forgery faces, selecting some common categories in forty types of LatTrans to edit the image’s facial attribute,  setting the pre-trained FaceSwapper by CelebA ~\cite{liu:deep} as the source image to replace stochastic face in the target image. Given an example as Fig.~\ref{fig1}(b), and record the forged image and mark the grounding bound box 
$y_{b o x}=\left\{x_{1}, y_{1}, x_{2}, y_{2}\right\}$ for detection and localization.
\nocite{karras:progressive}
\nocite{liu:deep}

\subsubsection{Real Image Fake text}
Text data exists multiple representations for the same semantic, which leads to insufficient compatibility pairs generated by text attribute editing or swap corresponding to traditional emotion reverse theory, and also unsuitable to combine the EFS images with simple textual person name replacement samples. Therefore, we through the state-of-the-art technology MiNiGPT-v2 ~\cite{chen:minigpt} and CogVLM ~\cite{wang:cogvlm} in Multimodal Large Language Models (MLLMs) ~\cite{tu:overview} fundamental both image and text content to produce forged text with equal probability, the similarity between image-text sample pairs is controlled to be 50$\%$ to 75$\%$ considering the noise in the real media news. Given an example as Fig.~\ref{fig1}(c), and recorded text labels for the detection.
\nocite{chen:minigpt}
\nocite{wang:cogvlm}
\nocite{tu:overview}

\subsubsection{Fake Image Fake Text}
To simulate the correlation between image and text in fake news, we randomly draw some sample pairs from the source pool manipulate the image part according to the above-mentioned means, and then combine the forged image as the proposed real image to generate forged text with MLLMs. Given an example as Fig.~\ref{fig1}(d), and record media news real or fake binary labels, forgery or origin image labels, authentic or falsehood text labels, and image manipulation localization labels.

\subsection{Data Statistics}
The overall statistics of the MFND dataset are illustrated in Fig.~\ref{fig2}(a). It contains 125k multimodal fake news samples, including 32869 real image real text news pairs, 40821 fake image real text news pairs, 29826 real image fake text news pairs, and 21484 fake image fake text news pairs. The dataset is divided into three parts, the 95k pairs are part of model training, another 15k pairs are part of testing and the remaining 15k are for testing.

We summarize the information of the major existing multimodal fake news datasets and ours in Fig.~\ref{fig2}(b). Earlier datasets such as Twitter ~\cite{khattar:mvae}, Pheme ~\cite{zubiaga：exploiting}, Weibo ~\cite{jin:multimodal} and NewsCLIPpings ~\cite{luo:newsclippings} without deepfake techniques and fake samples are generated by mismatched image and text pairs and contain only binary labels of the media news, recent DGM$^4$ ~\cite{shao:detecting} sets up image and text manipulation methods based on reversed sentiment transformations, containing both image and text localization annotations. MFND updates the technique grounded in real scenes and text semantic representations, utilizes more sophisticated deepfake methods in forged images exchanges sentiment-recognition-based text for generated text, provides rich fine-grained annotations, and removes redundant text localization.
\nocite{zubiaga：exploiting}
\nocite{luo:newsclippings}

\section{Method}
As shown in Fig.~\ref{fig3}, the overall framework of the proposed method Shallow-Deep Multitask Learning (SDML) contains shallow and deep inference stages. As shallow, we encode the image and text modality into a sequence of embeddings by two dedicated uni-modal Encoders, update the alignment results in uniform space under the momentum distillation-based Light Punishment Contrastive Learning module (LPCL), and learn the mutual modal features through Adaptive Cross-Modal Fusion module (ACMF) for predicting the Fake News Binary Detection outputs. Under deep, we design the two-branch framework to augment the uni-modal features for image and text separately, and eventually predict Image Forgery Detection and Localization and Text Forgery Detection with dedicated projections.

\subsection{Shallow Inference}
Given an input image-text pair $(V, T) \in D$, where $V$, $T$, and $D$ represent image, text, and dataset, we obtain the encoded image and text single modality features $V^{e}$ and $T^{e}$ by two pre-trained Encoders. The vision encoder chooses the 12-layer ViT-B/16 which has been initialized with ImageNet and adds a fully connected layer to dimension transform the encoded results. The text encoder chooses the first 8 layers of BERT-16 to extract the text semantics and also using a fully connected layer obtain the finally transformed text embedding.

\subsubsection{Light Punishment Contrastive Learning}

In multimodal studies, contrastive learning is generally used to embed two single modalities into a unified space thereby eliminating semantic differences. Contrastive learning adopts a unique one-hot encoding method thus all negative pairs will be penalized resulting in an inaccurate multimodal embedding. We propose utilizing knowledge distillation to obtain more views’ modality representation, combined with contrastive learning to soften the high punishment caused by weak correlations, such as mismatched image and text content in positive pairs, and effective matching of misaligned text in negative pairs.

Specifically, we put the encoded single modal embeddings $V^{e}$ and $T^{e}$ in two dedicated Multi-Layer Perceptron (MLP) then the features are transformed into the low-dimensional space obtain  $e^{v}$ and $e^{t}$, utilize the similarity function $(sim)$ measures similarity scores via dot product. Within a batch size $N$ of training, the predicted image-text similarity $p_{\mathrm{ij}}^{\mathrm{v} \rightarrow \mathrm{t}}$ and text-image similarity $p_{\mathrm{ij}}^{\mathrm{t} \rightarrow \mathrm{v}}$ are calculated as follows%
\setlength\abovedisplayskip{0.1cm}
\begin{align}
    p_{\mathrm{ij}}^{\mathrm{t} \rightarrow \mathrm{v}}=\frac{\exp \left(\operatorname{sim}\left(e_{i}^{v}, e_{j}^{t}\right) / \tau\right)}{\sum_{j=1}^{N} \exp \left(\operatorname{sim}\left(e_{i}^{v}, e_{j}^{t}\right) / \tau\right)},
\end{align}%
\begin{align}
    p_{\mathrm{ij}}^{t \rightarrow v}=\frac{\exp \left(\operatorname{sim}\left(e_{i}^{t}, e_{j}^{v}\right) / \tau\right)}{\sum_{j=1}^{N} \exp \left(\operatorname{sim}\left(e_{i}^{t}, e_{j}^{v}\right) / \tau\right)},
\end{align}%
where $\tau$ is the learnable temperature parameter with an initial value of 0.07, the results compose the similarity matrix $P$. At this point, the image-text contrastive loss $\mathcal{L}_{I T C}$ can be calculated by the mean of image-text and text-image loss, in which the two losses are both calculated as follows %
\begin{align}
    \mathcal{L}=-\frac{1}{N} \sum_{i=1}^{N} \sum_{j=1}^{N} y_{i j} \log p_{i j},
\end{align}% 
where the $y_{i j}$ is the corresponding one-hot vector to the real labels, in which the negative pair is denoted as 0 and the positive pair is denoted as 1.  

To balance the presence of weak correlations, we choose the unimodal exponential moving average version of the base model as the momentum model, the momentum features are denoted as $V^{s}$ and $T^{s}$, obtaining $s^{v}$ and $s^{t}$ and unifying the dimension with the above $e^{v}$ and $e^{t}$. We maintain two queues of size $M$ $(M\ll N)$ to store the most recent image-text pairs. The momentum image-text similarity $s_{\mathrm{ij}}^{v \rightarrow \mathrm{t}}$ and momentum text-image similarity $s_{\mathrm{ij}}^{t \rightarrow \mathrm{v}}$ are computed as follows %
\begin{align}
    s_{\mathrm{m}}^{\mathrm{t} \rightarrow \mathrm{v}}=\frac{\exp \left(\operatorname{sim}\left(s^{v}, s_{m}^{t}\right) / \tau\right)}{\sum_{m=1}^{M} \exp \left(\operatorname{sim}\left(s^{v}, s_{m}^{t}\right) / \tau\right)}.
\end{align}%
Results compose the momentum similarity matrix $S$. Momentum Multimodal loss $\mathcal{L}_{M M}$ is calculated by cross-entropy using the similarity matrix $P$ and momentum similarity matrix $S$. The final momentum distillation-based light punishment contrastive loss is calculated as follows, with $\lambda$ being the learnable parameter with an initial value of 0.02:%
\begin{align}
    \mathcal{L}_{L C}=\mathcal{L}_{I T C}+\lambda \mathcal{L}_{M M}.
\end{align}%

\subsubsection{Adaptive Cross-Modal Fusion}
Image and text as different modalities contain rich independent semantics and shared semantics. Existing research attempts to map image-to-text space or text-to-image, which ignores the unique qualities of discrete and continuous information resulting in semantic hidden. The activity of image and text features present higher and lower, thus needing to measure the importance as a shared part in modal fusion. We propose an adaptive cross-modal fusion module that can preserve the private characteristics of a single modality and adaptively adjust the mutual modal feature weights.

Specifically, given a sequence of image-text feature pairs, the previous layer vectors are denoted as $H_{l-1}$ where $l \in[1, L]$. We normalize the modalities method to the same dimension is calculated as follows %
\begin{align}
    {H}_{l-1}=L N_{V}(I)+L N_{T}(T),
\end{align}%
where $L N_{V}$ and $L N_{T}$ are the RMSNorm for image features and text features, respectively. Define the image-text attention operation using the key projection matrix and value projection matrix for images and the query projection matrix for text calculated as follows %
\begin{align}
    H_{l}^{Q_{0}}={H}_{l-1} W_{l}^{Q_{0}},
\end{align}%
\begin{align}
    H_{l}^{K_{1}}={H}_{l-1} W_{l}^{K_{1}}, H_{l}^{V_{1}}={H}_{l-1} W_{l}^{V_{1}},
\end{align}%
\begin{align}
    C_{l}^{0}=\operatorname{Softmax} \left(\frac{H_{l}^{Q_{0}} H_{l}^{K_{1} T}}{\sqrt{d}}\right) H_{l}^{V_{1}},
\end{align}%
where $W_{l}^{Q_{0}}$ $W_{l}^{K_{1}}$ $ W_{l}^{V_{1}}$ are the learnable projection matrix, $C_{l}^{0}$ is the image-text cross-context feature, as a local shared feature it contains the intersection part with the text while retains more image feature, in the same way, we obtain the text-image cross-context feature $C_{l}^{1}$. To adaptive handle mutual modality information, we use the learnable modal weight parameter $\alpha$, to sum up the local image-text cross-context feature and text-image cross-context feature, bring the result $C$ to pass through another attention mechanism again with the image-text sequence handle global fusion, add a linear aggregate layer final obtain mutual modality, which is calculated as follows %
\begin{align}
    C=(1-\alpha)C_{l}^{0}+\alpha C_{l}^{1},
\end{align}%
\begin{align}
    F=\text { Linear }\left(\text {Softmax} \left(\frac{H_{0} C^{T}}{\sqrt{D}}\right) C \right).
\end{align}%

To ensure modal collaboration in the same semantic space also used a shared Feed Forward Network (FFN). Mutual modality features $F$  are fed to a dedicated fake news detection projection to predict real and false news is denoted as $\tilde{y_{bin}}$, the media news binary loss is calculated as follows %
\begin{align}
    \mathcal{L}_{B I C}=-\left(y \log \tilde{y_{bin}}+\left(1-y\right) \log \left(1-\tilde{y_{bin}}\right)\right).
\end{align}%

\subsection{Deep inference}
\begin{table*}
    \setlength{\tabcolsep}{6pt}
    \setlength{\abovecaptionskip}{0.2cm}
    \centering
    \caption{Comparison of multi-modal multi-task models on MFND dataset.}
    \vspace{-0.5em}
    \begin{tabular}{cccccccccc}
        \toprule
        Categories  & \multicolumn{3}{c}{Multimodal} & \multicolumn{2}{c}{Image Binary} & \multicolumn{2}{c}{Image Grounding} & \multicolumn{2}{c}{Text Binary} \\
        \midrule
        Methods & AUC & ACC & mF1 & AUC & ACC & IoUmean & IoU50 & AUC & ACC \\
        \midrule
        CLIP & 79.53 & 73.48 & 73.09 & 80.54 & 72.16 & 46.04 & 46.04 & 69.61 & 65.46 \\
        ViLP & 82.52 & 77.43 & 77.58 & 82.88 & 75.51 & 50.39 & 54.99 & 71.13 & 68.61 \\
        DGM$^4$ & 91.62 & 84.92 & 84.73 & 92.59 & 85.47 & 74.32 & 81.25 & 94.02 & 92.71 \\
        RPPG-Fake & 91.42 & 84.22 & 84.67 & 90.11 & 83.45 & 70.26 & 79.59 & 90.47 & 88.44 \\
        \textbf{SDML (Ours)} & \textbf{92.43} & \textbf{85.54} & \textbf{85.83} & \textbf{95.65} & \textbf{88.41} & \textbf{77.83} & \textbf{84.39} & \textbf{95.76} & \textbf{93.10} \\
        \bottomrule
    \end{tabular}
     \vspace{-0.8em}
    \label{table1}
\end{table*}

\begin{table*}
    \setlength{\tabcolsep}{6pt}
    \setlength{\abovecaptionskip}{0.2cm}
    \centering
    \caption{Comparison of multi-modal multi-task models on DGM$^4$ dataset.}
     \vspace{-0.5em}
    \begin{tabular}{cccccccccc}
        \toprule
        Categories  & \multicolumn{3}{c}{Multimodal} & \multicolumn{2}{c}{Image Binary} & \multicolumn{2}{c}{Image Grounding} & \multicolumn{2}{c}{Text Binary} \\
        \midrule
        Methods & AUC & ACC & mF1 & AUC & ACC & IoUmean & IoU50 & AUC & ACC \\
        \midrule
        CLIP & 83.10 & 76.21 & 76.77 & 83.75 & 75.38 & 49.26 & 50.08 & 72.97 & 68.68 \\
        ViLP & 85.30 & 78.26 & 78. 44& 85.11 & 78.56 & 60.36 & 68.84 & 75.00 & 71.37 \\
        DGM$^4$ & 93.23 & \textbf{86.39} & 86.11 & 93.22 & 87.23 & 76.70 & 83.49 & 95.16 & 94.27 \\
        RPPG-Fake & 92.02 & 85.69 & 85.97 & 90.64 & 84.03 & 71.19 & 81.44 & 93.48 & 92.95 \\
        \textbf{SDML (Ours)} & \textbf{93.67} & 86.14 & \textbf{86.33} & \textbf{96.12} & \textbf{89.31} & \textbf{78.36} & \textbf{84.60} & \textbf{96.14} & \textbf{94.94} \\
        \bottomrule
    \end{tabular}
     \vspace{-0.9em}
    \label{table2}
\end{table*}

Mutual modality fuses semantically aligned image and text-sharing information, existing studies generally detect and locate forgery image and text in a single branch by mutual features. The instability of noise effects in real media news leads to semantic gaps between single modality and fusion modality while deepfake techniques exacerbate such semantic inaccuracies, thus it is difficult to predict all the forgery categories only through mutual modality features. The dual branch combines unimodal independent information, and deep reasoning for image and text to provide different views making the semantic gap complemented.

\subsubsection{Image Forgery Detection and Localization}
Image modality as visual perceptual information has an important impact on the media propagation degree, and manipulation techniques against image modality are sophisticated and complex and require fine-grained inference. To enhance the key visual to weaken other interferences, we utilize the advantages of two image encode methods namely combining multi-level and multi-scale visual perception to form a Multi-View Extractor (MVE).

Specifically, we adopt ViT-B/16 to model the long-range contextual information of the image by extracting different encoder layers, obtain multi-level visual intermediate feature mapping $\left\{I_{a}^{k}\right\}_{k=1}^{n}$, where $n$ denotes the number of layers, splice all the visual features along the channel dimensions to capture the subtle changes in the image. The multi-level visual perceptual features contain initial spatial information and later semantic indications, calculated as follows %
\begin{align}
    I_{a}^{1}=\operatorname{ViTBlock}_{1}(V),
\end{align}%
\begin{align}
   I_{a}^{k}=\operatorname{ViTBlock}_{k}\left(I_{a}^{k-1}\right), k=2, \ldots, n,
\end{align}%
\begin{align}
    I_{a}=\operatorname{Concat}\left[I_{a}^{1}, I_{a}^{2}, \ldots, I_{a}^{n}\right].
\end{align}%

To model the local information of the image, we choose CNN network which performs well in edge and texture processing, obtain multi-scale visual features $\left\{I_{b}^{k}\right\}_{k=1}^{m}$, where $m$ is the number of scales of the image including multiple resolutions such as $\frac{H}{4} * \frac{W}{4}$, $\frac{H}{8} * \frac{W}{8}$, $\frac{H}{16} * \frac{W}{16}$ where $H*W$ is the initial resolution, transform to a uniform channel dimension and connect all the extracted visual features. The multi-scale visual perceptual features contain spatial details at multiple grains, calculated as follows %
\begin{align}
    I_{b}^{1}=\operatorname{ConvBlock}_{1}(V_{1}),
\end{align}%
\begin{align}
   I_{b}^{k}=\operatorname{ConvBlock}_{k}\left(I_{b}^{k-1}\right), k=2, \ldots, m,
\end{align}%
\begin{align}
    I_{b}=\operatorname{Concat}\left[I_{b}^{1}, I_{b}^{2}, \ldots, I_{b}^{m}\right].
\end{align}%

Splice multi-level and multi-scale visual perceptual features by channel and align dimension with the mutual modality features by a linear layer to obtain independent image modality features, sum up it with mutual modality features $F$. The image deep inference features $I_{p}$ are fed to the dedicated detection and localization projection to predict the forgery or origin image and mark the forged bounding box, by cross-entropy loss for binary classification and smoothing L1 loss for bounding box regression, the total image loss is calculated as follows %
\begin{align}
    \mathcal{L}_{I M G}=\mathcal{L}_{I B \mathrm{ic}}(I_{p})+\mathcal{L}_{\text {IBox }}(I_{p}).
\end{align}%

\subsubsection{Text Forgery Detection}
The same semantic content can be represented in many different textual representations, thus the detection of text modality requires deep contextual interaction information. Pre-trained LLMs are trained on masses of language libraries to help them adapt to downstream tasks. Research in several areas has made significant progress based on this technology.

In this paper, we choose the last 4 layers of BERT-16 and a linear layer as the Context Aggregator (CA) and align dimension with mutual modality features $F$ to obtain a composite feature $T_{p}$ that fuses text uni-modal features and shared mutual modal features. Feed into the dedicated text detection projection to predict authentic or falsehood text binary classification is denoted as $\tilde{y_{t e x}}$, calculated as follows:
\begin{align}
    \mathcal{L}_{T E X}=-\left(y \log \tilde{y_{t e x}}+\left(1-y\right) \log \left(1-\tilde{y_{t e x}}\right)\right),
\end{align}%

The total loss $\mathcal{L}_{TOTAL}$ consists of four components, the light punishment contrastive loss $\mathcal{L}_{LC}$, the fake news loss $\mathcal{L}_{BIC}$, the image detection and localization loss $\mathcal{L}_{IMG}$ and the text detection loss $\mathcal{L}_{TEX}$,calculated as follows:%
\begin{align}
   \mathcal{L}_{TOTAL }=\mathcal{L}_{L C}+\mathcal{L}_{B I C}+\mathcal{L}_{I M G}+\mathcal{L}_{T E X}.
\end{align}%

\section{Experiments}
\subsection{Datasets, Metrics, and Settings}

The Twitter dataset ~\cite{khattar:mvae} consists of tweets containing textual information, visual information, and social contextual information related to them. The Weibo dataset ~\cite{jin:multimodal} is from Xinhua News Agency and Weibo where each tweet contains three elements i.e. tweet id, text, and image. To comprehensively evaluate the model, we use the accuracy (ACC), area under the receiver operating characteristic curve (AUC), and the mean of F1 scores (mF1) to evaluate fake news detection, AUC, and ACC values for image and text detection. To verify the bounding box prediction, image localization calculates the intersection ratio between the true and predicted coordinates(IoU), sets thresholds 0.5, 0.75, and 0.9 to calculate the average accuracy, and selects IoUmean and IoU50 as the evaluation scales.
\nocite{khattar:mvae}
\nocite{jin:multimodal}

For the multi-modal multi-task approaches, CLIP ~\cite{radford:learning} co-trains an image encoder and a text encoder to predict the correct pairing of a batch of image-text samples. ViLT ~\cite{kim:vilt}greatly simplifies the processing of visual inputs in the same convolution-free manner as text inputs. DGM$^4$ ~\cite{shao:detecting} fuses the methods of contrast learning, multimodal feature fusion Attentional Mechanisms. RPPG-Fake ~\cite{zhang:early} explores the problem through the generation of propagation paths. For multi-modal single-task approaches, HMCAN ~\cite{qian:hierarchical} uses a hierarchical contextual attention network, MEAN ~\cite{wei:modality} uses a multi-modal generator and a dual discriminator for adversarial training, COOLANT ~\cite{zhang:cross} uses an attention-guided contrast and cross-fertilization framework, and Event-Radar ~\cite{ma:event} uses a graph-structured event consistent encoder and multi-view fusion.
\nocite{radford:learning}
\nocite{kim:vilt}
\nocite{wei:modality}
\nocite{zhang:early}

All experiments are implemented on the Pytorch deep learning framework. The momentum queue size is set to 65535, the media detection, image detection, image localization, and text detection projections are set to three different multi-layer perceptual with output dimensions of 2, 2, 4, and 2. The model trained for 100 epochs with a batch size of 64, AdamW optimizer, with a weight decay of 0.005, in the first 1000 steps the learning rate is warmed up to $5 e^{-6}$, decaying to $5 e^{-7}$ after the cosine scheduling.

\begin{table}
    \setlength{\tabcolsep}{2pt}
    \setlength{\abovecaptionskip}{0.2cm}
    \centering
    \caption{Comparison with single-task models on three datasets.}
     \vspace{-0.5em}
    \begin{tabular}{ccccccc}
        \toprule
        Datasets  & \multicolumn{2}{c}{MFND} & \multicolumn{2}{c}{Weibo} & \multicolumn{2}{c}{Twitter} \\
        \midrule
        Methods  & ACC & mF1 & ACC & mF1 & ACC & mF1 \\
        \midrule
        HMCAN & 80.62 & 80.46 & 88.52 & 88.55 & 89.71 & 89.54 \\
        MEAN & 81.09 & 81.16 & 89.49 & 89.15 & 78.42 & 78.49 \\
        COOLANT & 83.11 & 83.63 & 92.30 & \textbf{92.63} & 90.04 & 90.81 \\
        Event-Randar & 82.68 & 82.89 & 91.94 & 91.90 & 92.84 & 92.31 \\
        \textbf{SDML (Ours)} & \textbf{84.29} & \textbf{84.31} & \textbf{92.61} & 92.44 & \textbf{93.07} & \textbf{93.61}\\
        \bottomrule
    \end{tabular}
     \vspace{-1.2em}
    \label{table3}
\end{table}

\subsection{Comparison Results}

\noindent\textbf{Results on MFND dataset.} 
We compare three multi-modal multi-task baseline methods with SDML on our newly proposed MFND dataset, and the results are shown in Table~\ref{table1}. SDML significantly outperforms the three baseline methods on all evaluation metrics, and the image localization task shows the most pronounced superiority, with IoUmean and IoU50 values exceeding the highest value by 3.51\% and 3.14\%, respectively. This suggests that shallow-deep inference under dual branching can comprehensively and accurately capture the interactions between images and text, and track the semantic changes caused by the operations.

\begin{table*}
    \setlength{\tabcolsep}{6pt}
    \setlength{\abovecaptionskip}{0.2cm}
    \centering
    \vspace{-0.8em}
    \caption{Ablation results of modalities and modules. `LPCL', `ACMF', `MVE', and `CA' are short for `Light Punishment Contrastive Learning', `Adaptive Cross-Modal Fusion',  `Multi-View Extractor', and `Context Aggregator', respectively.}
    \vspace{-0.4em}
    \begin{tabular}{cccccccccc}
        \toprule
        Categories  & \multicolumn{3}{c}{Multimodal} & \multicolumn{2}{c}{Image Binary} & \multicolumn{2}{c}{Image Grounding} & \multicolumn{2}{c}{Text Binary} \\
        \midrule
        Methods  & AUC & ACC & mF1 & AUC & ACC & IoUmean & IoU50 & AUC & ACC \\
        \midrule
        w/o Image & \quad & \quad & \quad & \quad & \quad & \quad & \quad & 77.62 & 73.77 \\
        w/o Text & \quad & \quad & \quad & 94.89 & 88.08 & 77.21 &83.92 & \quad& \quad \\
        w/o LPCL & 90.43 & 83.35 & 83.65 & 93.84 & 86.55 & 65.46 & 77.37 & 91.62 & 89.58 \\
        w/o ACMF & 89.06 & 82.19 & 81.94 & 92.70 & 85.37 & 64.08 & 74.09 & 91.73 & 89.43 \\
        w/o MVE & 91.25 & 84.15 & 84.13 & 93.38 & 86.25 & 74.16 & 82.75 & 93.86 & 91.27 \\
        w/o CA & 91.44 & 84.92 & 84.49 & 94.77 & 87.71 & 76.49 & \textbf{84.47} & 93.29 & 91.09 \\
        \textbf{SDML (Ours)} & \textbf{92.43} & \textbf{85.54} & \textbf{85.83} & \textbf{95.65} & \textbf{88.41} & \textbf{77.83} & 84.39 & \textbf{95.76} & \textbf{93.10} \\
        \bottomrule
    \end{tabular}
    \vspace{-0.8em}
    \label{table4}
\end{table*}

\vspace{0.2em}
\noindent\textbf{Results on DGM$^4$ dataset.} 
We compare our methods with the same multi-modal multi-task methods on the DGM$^4$ dataset, and the comparison results are listed in Table~\ref{table2}. The results show that our method outperforms DGM$^4$ by 2\% in both AUC and ACC values on the image detection task, which indicates that unimodal data enhancement guides feature representation learning. SDML also performs superiorly on multimodal and text detection tasks, e.g., the AUC value of multimodal is improved by 0.44\% and the AUC value of text is enhanced by 0.98\%.

\vspace{0.2em}
\noindent\textbf{Comparison with single-task methods.}
We also evaluate our method as well as four multi-modal single-task methods on three datasets, all of which perform binary detection of multimodal fake news. The comparison results are shown in Table~\ref{table3}, where the ACC and F1 values of our models are significantly better than the other baseline methods on all datasets. The predicted values of all methods on the MFND dataset are lower than those on the microblogging and Twitter datasets, which suggests that deepfake data has high manipulation complexity and needs to be combined with different view information from unimodal and multimodal to mine deep semantics. In addition, comparing the different performances of SDML on the same dataset, the effect of binary classification under single-task decreases significantly, which suggests that multi-task guides the model to learn stable features to improve the prediction accuracy.

\begin{figure}[!t]
    \centering
    \setlength{\abovecaptionskip}{0.2cm}
    \begin{minipage}[b]{0.48\linewidth}
        \centering
        {\includegraphics{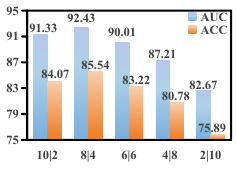}}%
        \vspace{-0.5em}
        \caption{Ablation on layer numbers of text encoder and Contextual Aggregator.}
        \label{fig4}
    \end{minipage}
    \hfill
    \begin{minipage}[b]{0.48\linewidth}
        \centering
        {\includegraphics{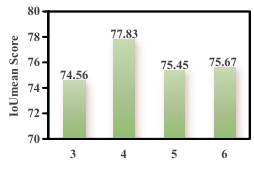}}%
        \vspace{-0.5em}
        \caption{Ablation on numbers of Multi-view Extractor.}
        \label{fig5}
    \end{minipage}
    \vspace{-0.9em}
\end{figure}

\subsection{Ablation Study}
\textbf{Ablation of modalities.} To validate the multimodal relevance of images and text, we perform an ablation experiment, and the results from rows 1-2 of Table~\ref{table4} show that the full multimodal version outperforms the eliminated portion. The experiments demonstrate that multimodal information interacts with each other and capturing common semantics is particularly important for feature learning. 

\vspace{0.2em}
\noindent\textbf{Ablation of different modules.}
To verify the importance of different modules to our SDML model, we set up several ablation experiments, and the results are listed in rows 3-6 of Table~\ref{table4}. We can find that removing light punishment contrastive learning (row 3) or removing adaptive cross-modal fusion (row 4) results in a sharp decrease in accuracy for each task, suggesting that the model needs to complete modal alignment and fusion at an early stage to facilitate deeper reasoning. In addition, removing the multi-view extractor (row 5) and context aggregator (row 6) is correspondingly less effective, suggesting that the unimodal semantics contain separate information from the mutual modality and can be used as complementary augmentation data. 

\vspace{0.2em}
\noindent\textbf{Impact of layer numbers of text encoder and Contextual Aggregator.} We consider five scenarios regarding the layers for the text encoder and contextual aggregator. The settings and results are shown in Fig.~\ref{fig4}. In terms of the evaluation metrics AUC and ACC, models with ratio setting 8$|$4 (i.e., 8 layers text encoder and 4 layers contextual aggregator) obtain the best results while performance degrades faster when the ratio becomes larger. 

\vspace{0.2em}
\noindent\textbf{Impact of numbers of Multi-view Extractor.} The multi-view extractor is composed of multi-layer and multi-scale visual perception modules. Here we compare layers number 3, 4, 5, and 6 in combination with the corresponding scale number, and select the IoUmean as the evaluation score. The results in Fig.~\ref{fig5}, illustrate that the best performance can be achieved when the values of n and m are set to be 4.

\begin{figure}[!t]
    \setlength{\abovecaptionskip}{0.2cm}
    \centering
    \includegraphics{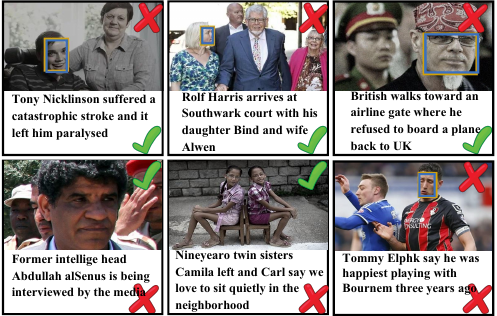}
    \vspace{-0.3em}
    \caption{Visualization of our predictions (real in green Checkmark while fake in red XSolid). The ground truth and prediction bounding boxes in the images are in yellow and blue, respectively. } 
    \label{fig6}
    \vspace{-0.8em}
\end{figure}

\subsection{Visualization and Analysis}
We provide some visualization results for fake news detection and localization in Figure~\ref {fig6}. The first row is all fake news in the Fake Images Real Text category, generated by FS, EFS, and AM from left to right, and the second row is all fake news in the Fake Text category, in which the faked images of the rightmost news samples are combined and the rest of the samples are real images. The visualization results use pairwise error symbols to mark the detection results and use bounding boxes to locate the forgery images, which demonstrates that our method can accurately detect manipulated images under different types of deep forgery techniques, as well as manipulated text generated by a large language model under the influence of real media news noise.

\section{Conclusion}
In this paper, we propose a large-scale complex MFND dataset with richer annotations to forge multimodal news through the latest generation techniques. To facilitate the detection and localization tasks in fake news detection, we propose a Shallow-Deep Multitask Learning (SDML) model to accomplish a deeper understanding of the interactions between unimodal and mixed modalities, improving its interpretability and robustness. We evaluate its effectiveness and superiority by conducting baseline comparison synthesis experiments on mainstream and MFND datasets.

\section*{Acknowledgments}
This work was supported by Natural Science Foundation of Hebei Province under the Grant F2024202017, Basic Research Project of Hebei Universities in Shijiazhuang under the Grant 241790817A, the Central Guidance Fund for Local Science and Technology Development Projects under Grant 246Z0106G, Guangdong Basic and Applied Basic Research Foundation (Grant No. 2023A1515140037), Guangdong Provincial Key Laboratory (Grant 2023B1212060076), and Guangdong Key Laboratory of Information Security Technology, Sun Yat-sen University. The computational resources are supported by SongShan Lake HPC Center (SSL-HPC) in Great Bay University.

\bibliographystyle{named}
\bibliography{main}

\end{document}